\documentclass{article}

\usepackage{arxiv}

\usepackage[utf8]{inputenc} \usepackage[T1]{fontenc}    
\PassOptionsToPackage{hyphens}{url}

\usepackage[breaklinks,backref=page]{hyperref}

\usepackage{url}            \usepackage{booktabs}       \usepackage{amsmath}        \usepackage{amsfonts}       \usepackage{amssymb}
\usepackage{nicefrac}       \usepackage{microtype}      \usepackage{lipsum}         \usepackage{graphicx}
\usepackage[numbers,sort&compress]{natbib}
\usepackage{doi}
\usepackage{float}
\usepackage{multirow}
\usepackage{lineno}

\usepackage{cleveref}

\usepackage{wrapfig}

\usepackage{authblk}

\usepackage[justification=centering]{caption}

\usepackage[inline]{enumitem}

\usepackage{bbding}

\usepackage[flushleft]{threeparttable}

\usepackage{listings}

\usepackage{subcaption}

\usepackage{orcidlink}

\usepackage{bm}

\renewcommand*{\backref}[1]{}
\renewcommand*{\backrefalt}[4]{
  \ifcase #1
    No citations.\or
(Cited on page: #4).
  \else
(Cited on pages: #4).
  \fi
}

\newcommand{\appendixref}[1]{\hyperref[#1]{Appendix~\ref{#1}}}

\title{Play Like Champions: Counterfactual Feedback Generation in Latent Space }

\date{}	

\makeatletter
\newcommand\email[2][]{\newaffiltrue\let\AB@blk@and\AB@pand
      \if\relax#1\relax\def\AB@note{\AB@thenote}\else\def\AB@note{\relax}
        \setcounter{Maxaffil}{0}\fi
      \begingroup
        \let\protect\@unexpandable@protect
        \def\thanks{\protect\thanks}\def\footnote{\protect\footnote}
        \@temptokena=\expandafter{\AB@authors}
        {\def\\{\protect\\\protect\Affilfont}\xdef\AB@temp{#2}}
         \xdef\AB@authors{\the\@temptokena\AB@las\AB@au@str
         \protect\\[\affilsep]\protect\Affilfont\AB@temp}
         \gdef\AB@las{}\gdef\AB@au@str{}
        {\def\\{, \ignorespaces}\xdef\AB@temp{#2}}
        \@temptokena=\expandafter{\AB@affillist}
        \xdef\AB@affillist{\the\@temptokena \AB@affilsep
          \AB@affilnote{}\protect\Affilfont\AB@temp}
      \endgroup
       \let\AB@affilsep\AB@affilsepx
}
\makeatother

\author[1]{\textbf{Andrzej Białecki}\protect\footnotemark[1]\hspace{0.5em}\textsuperscript{,}}
\affil[1]{Warsaw University of Technology}

\author[2]{\textbf{Adam Mastalerz}\protect\footnotemark[1]\hspace{0.5em}\textsuperscript{,}}
\affil[2]{Silesian University of Technology}

\author[3]{\textbf{Han Zhou}\protect\footnotemark[1]\hspace{0.5em}\textsuperscript{,}}
\affil[3]{University of British Columbia}

\hypersetup{
pdftitle={Play Like Champions: Counterfactual Feedback Generation in Latent Space},
pdfauthor={Andrzej Białecki, Adam Mastalerz, Han Zhou},
pdfkeywords={generative artificial intelligence, latent space traversal, optimal transport, variational autoencoder, esports},
}

\providecommand{\parencite}[1]{}

\renewcommand{\parencite}[1]{\cite{#1}}

\begin{document}
\renewcommand{\figureautorefname}{Fig.}
\renewcommand{\subsectionautorefname}{Subsection}

\maketitle

\footnotetext[1]{contact (in-order): \texttt{andrzej.bialecki94@gmail.com}, \texttt{adam.mastalerz@polsl.pl}, \texttt{hzhou30@student.ubc.ca}}

\begin{abstract}

    Recent advances in reinforcement learning have produced superhuman agents across a wide range of competitive games. As a byproduct, researchers have begun studying how these agents play, extracting behavioral representations, analyzing decision structure, and modeling the latent geometry of expert performance. However, this growing body of work has overwhelmingly focused on defeating human players rather than providing feedback, leaving a critical gap in creating model solutions to improve human players. Unlike chess and Go, where AI has become integral to player training, real-time strategy (RTS) games lack principled frameworks for translating expert knowledge into actionable feedback. We introduce Latent Maps of Performance, a framework for counterfactual path generation. We focus on StarCraft~II data to model player improvement as an algorithmic recourse within a learned representation space. As inspiration for our work, we have looked at the championship model used in sports science. We trained a Guided Variational Autoencoder model on 23,305 professional tournament replays, with macro-economic gameplay structure conditioned on match outcome, enabling counterfactual traversal between losing and winning gameplay profiles. To fulfill our goal, we have devised and verified four traversal strategies on out-of-distribution (OOD) data randomly sampled from a dataset of amateur replays, namely linear interpolation, iterative optimal transport, density-regularized gradient ascent, and neural flow matching, each designed to generate multi-step improvement trajectories that remain grounded in observed expert behavior while moving a player's profile toward winning configurations. Feedback is extracted at multiple granularities to support players at different stages of improvement. Finally, we conclude that there is a trade-off between the path-finding methods we employ and hope that future research will focus on developing model solutions for human improvement. 
\end{abstract}


\section{Introduction}
\label{sec:introduction}

Mastering real-time strategy (RTS) games such as StarCraft~II has long stood as a grand challenge for artificial intelligence, one only partially addressed by recent advances in reinforcement learning (RL)~\cite{Vinyals2017StarCraftChallenge, Mathieu2023AlphaStarUL}. The difficulty stems from the cognitive demands these games place on their players: precise control of units, careful management of economies, and continuous decision-making in adversarial settings where even momentary lapses can prove decisive. Success thus hinges on the interplay of multitasking, strategic foresight, and rapid reaction~\cite{Thompson2013}, making RTS an especially rich testbed for studying intelligent behavior. These high-frequency interactions make RTS games especially well-suited for large-scale behavioral study through open-source replay parsers and direct game-engine access~\cite{Bialecki2023SC2EGSet, Ferenczi2024}.
A growing body of work leverages data to surface game information for player decision support, both through digital interfaces and physical prototypes~\cite{Rijnders2022}. In StarCraft~II, community tools such as sc2replaystats~\cite{URLSc2replaystats} and replayman~\cite{URLReplaymanStego} have emerged to support replay analysis, alongside real-time dashboards that contextualize gameplay for spectators and post-match review~\cite{Charleer2018}. Strategic summaries and encounter-level analysis are highly valued by players across genres~\cite{Wallner2016WoT}. Game state retrieval by similarity to estimate win probabilities on demand is a possibility~\cite{Xenopoulos2022ggViz}. In parallel, AI methods have become deeply embedded in game research and development, powering procedural content generation~\cite{Shaker2016}, voice-driven agents that deepen immersion~\cite{Wei2025FACUL}, human-like behavior modeling~\cite{Sestini2025}, and automated quality assurance~\cite{Tufano2022}. However, most existing analysis tools remain oriented toward broadcast and streaming audiences rather than the players themselves~\cite{Kokkinakis2020DAX}.

This player-facing gap is not unique to gaming. In robotics, efficient simulators have driven dramatic breakthroughs~\cite{NVIDIANewton2025, MittalIsaacLab2025}, producing systems that now rival or exceed human performance in domains as varied as drone racing~\cite{Kaufmann2023, Lamberti2024}, badminton~\cite{Ma2025, Liu2026}, and table tennis~\cite{Durr2026}. Such interdisciplinary efforts are increasingly recognized as accelerators of research progress~\cite{Leite2025}. However, despite agents and robotic systems consistently surpassing average human ability, few of these works offer mechanisms to translate the resulting expertise back to human practitioners seeking to improve. The skill translation problem is well understood in the sport sciences, where the championship model describes how athletes shorten the path to performance gains by adopting the training methods, techniques, and tactics of successful peers~\cite{Hancock2011,Sozanski2015}. In domains where the competitive space is naturally digitized, this dynamic increasingly extends to AI. In chess, AI has reshaped human learning by serving as a scalable training partner. Analyses show that elite human play has steadily improved across the engine era~\cite{McIlroyYoung2020Chess, Gaessler2023Chess, Bilalic2026Chess}. AlphaZero's games further illustrate how superhuman agents can surface novel strategic ideas for human study~\cite{Sadler2019}. Similar patterns have emerged in Go following the rise of superhuman agents~\cite{Kang2022,Shin2021}.

RTS games share the same computational substrate, but, to our knowledge, no comparable bridge exists between agent expertise, representational learning, and human improvement. In this work, we introduce a latent-space feedback system that learns compressed representations of quantitative gameplay features from StarCraft~II replays and provides manifold-aware improvement guidance to players. Our contribution is a framework for training representational models that recover counterfactual improvement trajectories and reconstruct them back into the original feature space. Given a well-trained model, points sampled along an ``improvement trajectory'' in latent space can be decoded and compared with the player's input vector, yielding immediate feedback on which gameplay features need to change to improve performance, as determined by the model. Our work builds directly on ``SC2EGSet'', rather than asking ``\emph{who will win}'', we ask ``\emph{what the player should do differently}''.

\section{Related Work}
\label{sec:related_work}

Our work sits at the intersection of four research threads: StarCraft~II as a machine learning domain; variational autoencoders and disentangled representation learning; latent space traversal; and counterfactual explanations as actionable feedback. We discuss each in turn and position our contribution relative to prior art.

\paragraph{StarCraft~II as a Machine Learning Domain:} StarCraft~II has become a canonical benchmark for sequential decision-making under partial observability. \citet{Vinyals2019AlphaStar} demonstrated that a combination of imitation learning, multi-agent self-play, and a latent conditioning variable for strategy style can produce grandmaster-level play, establishing that large-scale replay data contains rich, learnable structure. However, AlphaStar is an autonomous agent; it optimizes for winning, not for explaining to human players how to improve. On the other hand, the simplistic nature of benchmarks geared primarily towards multi-agent solutions does not fit well in the context of providing feedback to players~\cite{samvelyan2019SMAC,ellis2023smacv2}. Work on modeling player skill from replays predates AlphaStar. \citet{Avontuur2013SkillModeling} showed that even simple classifiers trained on APM and economy features can predict a player's league with meaningful accuracy. Subsequent work demonstrated that macro-level economic measures, such as the Spending Quotient introduced by \citet{Bowman2021WinPrediction}, are among the strongest predictors of both skill and match outcomes.

\paragraph{Variational Autoencoders and Disentangled Representations:} Naturally, the Variational Autoencoder (VAE)~\cite{Kingma2022VAE} acts as the backbone for our work. By learning a probabilistic encoder and decoder jointly with a Kullback-Leibler (KL) divergence regulariser, the VAE produces a smooth, continuous latent space from which new samples can be reconstructed. As an extension, \citet{Higgins2017betaVAE} introduced $\beta$-VAE, and addressed the interpretability by up-weighting the KL term. The $\beta$ factor was applied to force the model to trade reconstruction fidelity for statistical independence between latent dimensions. Guided VAE addresses this in a direct manner~\citet{Ding2020GuidedVAE}. It attaches a supervised classifier to designated latent dimensions, and uses an adversarial excitation-inhibition mechanism to concentrate the target factor in that dimension while preventing it from leaking into the remaining dimensions. \citet{Schrum2025SAIL} presented ``SAIL'', which learns persistent skill embeddings from naturalistic behavioral data using expert-novice basis blending and counterfactual subskill swaps applied to motor tasks such as driving and baseball batting. We draw inspiration and intuition from these works.

\paragraph{Latent Space Traversal:} Generating semantically meaningful paths through a learned latent space is a non-trivial problem. Naive linear interpolation between two latent codes can pass through low-density regions of the prior, leading to decoded samples that lie outside the data manifold. \citet{Korkmaz2018OTMaps} formalize this distribution mismatch and show that optimal transport (OT) maps can correct linear trajectories so that all intermediate points remain consistent with the prior distribution while minimally deviating from a straight line. \citet{Song2023LatentTraversals} propose a more general framework that models latent structures as learned dynamic potential landscapes, deriving traversal trajectories as the gradient flow of a partial differential equation (PDE)-based potential field. \citet{Yeh2023OutcomeGuidedCF} take a related approach in the explainability domain with partial focus on StarCraft~II by leveraging a fixed ``SC2 Assault'' scenario. They generate counterfactuals from a jointly trained generative latent space where the traversal is guided toward a target outcome during decoding.

\paragraph{Counterfactual Explanations as Actionable Feedback:} Counterfactual explanations answer the question: ``\emph{what is the minimal change to the input that would change the model's prediction?}'' In a performance-improvement context, this is equivalent to algorithmic recourse. Providing a ranked list of feature changes that move a player from their current state to a more desirable one. \citet{Crupi2022CEILS} proposed ``CEILS'', generating counterfactuals as interventions in the latent space of a trained VAE~\cite{Crupi2022CEILS}. Work beyond passive explanation towards actionable coaching was shown by \citet{Bae2025DriverTraining}, leveraging counterfactual explanations in racing scenarios with language-based guidance. \citet{Pegios2024RiemannianCF} extend latent-space counterfactual generation by equipping the VAE latent space with a Riemannian metric pulled back through both the decoder and the classifier. 
\section{Material and Methods}
\label{sec:material_and_methods}

\paragraph{Replay Preprocessing and Feature Extraction:} To fulfill our goal of providing feedback based on learned representations, we have decided to use a dataset consisting of professional StarCraft~II games named ``SC2EGSet''~\cite{Bialecki2023SC2EGSet} licensed under CC-BY 4.0. Please refer to \appendixref{sec:sc2_description} for a simplified game description. At the time our work was prepared, the dataset consisted of 23,476 files containing game-state information sourced from 71 ``replaypacks''. Before training, each replay is converted into a tensor containing information about both players. For every player, we extract a 196-dimensional feature vector $\mathbb{R}^{196}$. Selected features primarily include economical game progression statistics that inform various aspects of the game. Additionally, for more expressive modeling insights, we split the array of player statistics into three windows, each spanning one-third of the total game duration. These features are named as ``early'', ``middle'', and ``late game'' windows ($3 \cdot 39$ features). Tracker statistics averaged within each window. Finally, we include the final economy state ($39$ features) and the economy difference features ($39$ features), computed as late-game economy minus early-game economy. Besides the economy, we include a scalar value, ``supply capped percent'', denoting the percentage of the game duration during which the player was unable to build additional units due to insufficient in-game infrastructure. The final replay tensor has shape $\mathbb{R}^{2 \times 196}$, where the first dimension corresponds to the player. The label used for disentanglement is the match result from player 0's perspective, represented as a binary outcome $y \in {0, 1}$, where $y = 1$ indicates that player 0 won. Prior to training, all features are standardized using per-feature z-score normalization. The normalization statistics (mean and standard deviation) are computed exclusively on the training set and subsequently applied to the validation and test sets, preventing any data leakage. A small epsilon ($\epsilon = 10^{-8}$) is added to each standard deviation to avoid division by zero for constant features.

The initial split between training, validation, and test sets was random (80\%/10\%/10\%). Replays with missing player statistics, player information, or undecided/draw outcomes were skipped. Finally, to assure that the training, and validation sets were indeed (80\%/10\%), the missing samples were taken out of the test set without replacement. After excluding samples, the splits consisted of the following numbers of samples in the training set ($n_{training} = 18780$), the validation set ($n_{validation} = 2347$), and the test set ($n_{test} = 2178$). To confirm the efficacy of our method on out-of-distribution (OOD) data, we have randomly sampled an additional ($n_{ood} = 2178$) samples from an unreleased dataset of players who submitted their replays to the sc2replaystats~\cite{URLSc2replaystats} in 2016-2020. Access to all of the pre-processed data and code is available. For more information please see~\appendixref{sec:appendix_data_code}.

\paragraph{Modelling}

To ensure the possibility of transitioning between learned latent space representations and reconstruction of the player features, we have decided to use a model adapted from the original Guided VAE~\cite{Ding2020GuidedVAE} architecture. We guide the latent space separation by the game outcome. Therefore, a part of the latent representation is encouraged to contain information about features that are significant for predicting winning or losing outcomes. The structure of our model consists of a symmetrical encoder-decoder multilayer perceptron (MLP) network with ReLU activations, with an additional supervised classifier attached to the latent space. In our modified Guided VAE~\cite{Ding2020GuidedVAE}, a selected number of dimensions are set to be supervised. An adversarial classifier is trained on the remaining latent dimensions, excluding the supervised dimensions. In training, the supervised classifier receives concatenated supervised latent dimensions of both players, $[z^{(0)}{1:k} | z^{(1)}{1:k}]$, and is trained to predict $y$ directly. At inference-time, when generating improvement paths for the player in seat 1, the input order is swapped, and the output probability is complemented, so that the score always represents the win probability of the player whose path is being improved. Please refer to~\appendixref{sec:model_details} for more implementation details. We use AdamW optimizers~\cite{Kingma2017adam,Loshchilov2018AdamW}, implemented in PyTorch. The first updates the VAE and its guided classifier jointly. The second trains an auxiliary adversarial classifier on the non-guided (free) latent dimensions. The third updates the VAE parameters adversarially, penalising the encoder for producing free-dimension representations that are predictive of match outcome.  Validation is monitored using the VAE loss, and early stopping is used to avoid overfitting. We conducted a hyperparameter search to find the best-performing model, see~\appendixref{sec:hparams}. All of the experiments were run on consumer hardware, see~\appendixref{sec:hardware}.

\paragraph{Latent Space Traversal and Feedback Generation:} After training, a replay is encoded into a latent vector. We construct a path from the current game state to a winning region in the supervised dimensions of the latent space. Each point interpolated along the path in the latent space can be reconstructed back to the feature space, where the difference between the current feature values and the reconstructed feature values informs the player about the improvement ``trajectory''. We compute feedback in three ways. First, we measure the raw change from the start of the path to the end. Second, we compute a minimum viable change that stops at the first waypoint where the predicted win probability crosses 0.5, if such a crossing exists. Third, we compute a win-probability-weighted change, where each step's feature change is weighted by the corresponding increase in predicted win probability; steps where win probability decreases contribute zero weight. The final output is a ranked list of feature changes. These changes act as rough suggestions, and should be interpreted as model-generated hypotheses. Example generated feedback reports are available in \appendixref{sec:example_feedback}.

\section{Standardized Path Representation}

To ensure compatibility with downstream generation and visualization tasks, the output of the path charting pipeline is strictly standardized, regardless of the specific employed strategy. The generated path $\mathcal{P}$ is defined as an ordered sequence of $n$ discrete waypoints in the $d$-dimensional latent space. This sequence is structured as a matrix $\mathbf{P} \in \mathbb{R}^{n \times d}$, where each row $\mathbf{p}_j \in \mathbb{R}^d$ corresponds to a specific waypoint along as seen in \autoref{eq:path_representation}.

\begin{equation}
    \mathbf{P} = 
    \begin{bmatrix}
    \text{---} & \mathbf{p}_0 & \text{---} \\
    \text{---} & \mathbf{p}_1 & \text{---} \\
    \vdots & \vdots & \vdots \\
    \text{---} & \mathbf{p}_{n-1} & \text{---}
    \end{bmatrix}
    =
    \begin{bmatrix}
    \text{---} & \mathbf{z}_{start} & \text{---} \\
    & \vdots & \\
    \text{---} & \mathbf{z}_{end} & \text{---}
    \end{bmatrix}
    \label{eq:path_representation}
\end{equation}
By definition across all strategies, the first waypoint $\mathbf{p}_0$ corresponds exactly to the initial latent vector $\mathbf{z}_{start}$. The final waypoint $\mathbf{p}_{n-1}$ corresponds to the terminal state of the strategy, denoted generally as $\mathbf{z}_{end}$ (which may represent a predefined target $\mathbf{z}_{target}$, a successfully converged optimization state, or the integrated endpoint of a velocity field). Traversal methods have their own set of tunable parameters fully described in~\appendixref{sec:hparams}.

\subsection{Linear Strategy}

The linear strategy serves as the foundational baseline for counterfactual generation. It implements a simple Linear Interpolation (LERP) to navigate the high-dimensional latent space, constructing a straight-line trajectory between the initial losing latent vector and a specified winning target. \textbf{Target Selection Strategies:}~Because the "winning" state is represented by a distribution of points rather than a single vector, the algorithm first collapses the winning latents $\mathcal{Z}_{win} = \{\mathbf{w}_1, \dots, \mathbf{w}_M\}$ into a single target vector $\mathbf{z}_{target} \in \mathbb{R}^d$ using one of two selectable methods. \textbf{Centroid Strategy (\texttt{method="centroid"}):} The target is defined globally as the unweighted mean position (centroid) of all known winning latent vectors. This provides a robust, generalized direction toward the center of the winning class, see \autoref{eq:target_centroid}. \textbf{Nearest Neighbors Strategy (\texttt{method="nearest"}):} To preserve local manifold structure and find the "closest" way to win, the target is derived locally. It calculates the mean of only the $k$ winning latents that are closest to the starting sample $\mathbf{z}_{start}$ (where $k$ is defined by \texttt{k\_neighbours}), see \autoref{eq:target_nearest}.

\noindent
\begin{minipage}{0.5\textwidth}
\begin{equation}
    \mathbf{z}_{target} = \frac{1}{M} \sum_{i=1}^{M} \mathbf{w}_i
    \label{eq:target_centroid}
\end{equation}
\end{minipage}\begin{minipage}{0.5\textwidth}
\begin{equation}
    \mathbf{z}_{target} = \frac{1}{k} \sum_{\mathbf{w} \in \text{k-NN}(\mathbf{z}_{start}, \mathcal{Z}_{win})} \mathbf{w}
    \label{eq:target_nearest}
\end{equation}
\end{minipage}
\vspace{\belowdisplayskip}

\textbf{Linear Interpolation (LERP) Formula:}~Once the target $\mathbf{z}_{target}$ is established, the path is generated as a sequence of $n$ waypoints $\{ \mathbf{p}_0, \mathbf{p}_1, \dots, \mathbf{p}_{n-1} \}$. The interpolation coefficient $\alpha_j$ is defined by an evenly spaced linear progression from $0.0$ to $1.0$, see \autoref{eq:interpolation_coefficient_lerp}. Each intermediate point $\mathbf{p}_j$ along the trajectory is calculated using a convex combination of the start and target vectors, moving progressively closer to the target as $\alpha$ increases, see \autoref{eq:intermediate_points_linear}. This straight-line interpolation assumes a globally Euclidean latent space, transitioning semantic features at a constant velocity without explicit regard for the underlying data density.

\noindent
\begin{minipage}{0.55\textwidth}
\begin{equation}
    \alpha_j = \frac{j}{n - 1}, \quad \text{for } j \in \{0, 1, \dots, n-1\}
    \label{eq:interpolation_coefficient_lerp}
\end{equation}
\end{minipage}\begin{minipage}{0.45\textwidth}
\begin{equation}
    \mathbf{p}_j = (1 - \alpha_j) \mathbf{z}_{start} + \alpha_j \mathbf{z}_{target}
    \label{eq:intermediate_points_linear}
\end{equation}
\end{minipage}

\subsection{Iterative Optimal Transport Strategy}

The iterative optimal transport (OT) strategy simulates a vector field flow toward the winning distribution. Instead of interpolating toward a single static target (such as a centroid), the algorithm dynamically recalculates a local barycentric target at each step by finding the optimal mass transport plan between the current position and the target distribution. \textbf{Distance Matrix and Transport Plan:}~At each step $t$, the current latent position $\mathbf{z}^{(t)}$ is treated as a point mass with weight $\mathbf{a} = 1$. The set of $N$ winning latents $\mathcal{Z}_{win} = \{\mathbf{w}_1, \dots, \mathbf{w}_N\}$ is treated as a uniform target distribution with weights $\mathbf{b} = \frac{1}{N} \mathbf{1}$. First, the squared Euclidean distance matrix $\mathbf{M}^{(t)} \in \mathbb{R}^{1 \times N}$ is computed. To prevent numerical underflow during exponentiation in the regularized transport step, the distance matrix is normalized by its maximum value as seen in \autoref{eq:norm_dist_matrix}. Next, the optimal transport plan $\mathbf{T}^{(t)}$ is computed. If entropic regularization $\lambda > 0$ (\texttt{ot\_reg}) is provided, the mathematically stable log-domain Sinkhorn algorithm is utilized as seen in \autoref{eq:ot_plan}, where $H(\mathbf{T})$ is the entropy of the coupling matrix. If $\lambda = 0$, the exact Earth Mover's Distance (EMD) is computed. \textbf{Local Barycentric Target:}~The resulting transport plan $\mathbf{T}^{(t)}$ dictates the optimal distribution of mass from the current position to the winning points. These transport weights are normalized to form a localized probability distribution as seen in \autoref{eq:norm_weights}. The local target $\mathbf{z}_{target}^{(t)}$ is then defined as the barycenter (weighted average) of the winning points, pulled specifically according to the optimal transport plan presented as \autoref{eq:target_barycenter}. \textbf{Euler Step (Vector Flow):}~Rather than jumping directly to the target, the algorithm treats the vector $(\mathbf{z}_{target}^{(t)} - \mathbf{z}^{(t)})$ as a local velocity field. It takes a small Euler step of size $\eta$ (\texttt{step\_size}) toward the local barycenter, as seen in \autoref{eq:vector_flow}. This process is repeated iteratively to trace a smooth trajectory. Because the transport plan dynamically updates at each spatial step, the resulting path closely mimics a continuous vector flow into the densest regions of the winning distribution.

\vspace{\abovedisplayskip}
\noindent
\begin{minipage}{0.5\textwidth}
\begin{equation}
    M_i^{(t)} = \frac{\|\mathbf{z}^{(t)} - \mathbf{w}_i\|^2}{\max_j \|\mathbf{z}^{(t)} - \mathbf{w}_j\|^2 + \epsilon}
    \label{eq:norm_dist_matrix}
\end{equation}
\end{minipage}\begin{minipage}{0.5\textwidth}
\begin{equation}
    \mathbf{T}^{(t)} = \arg\min_{\mathbf{T} \in \Pi(\mathbf{a}, \mathbf{b})} \langle \mathbf{T}, \mathbf{M}^{(t)} \rangle - \lambda H(\mathbf{T})
    \label{eq:ot_plan}
\end{equation}
\end{minipage}
\vspace{\belowdisplayskip}
\vspace{\abovedisplayskip}
\noindent
\begin{minipage}{0.5\textwidth}
\begin{equation}
    \tau_i = \frac{T_i^{(t)}}{\sum_{j=1}^N T_j^{(t)} + \epsilon}
    \label{eq:norm_weights}
\end{equation}
\end{minipage}\begin{minipage}{0.5\textwidth}
\begin{equation}
    \mathbf{z}_{target}^{(t)} = \sum_{i=1}^{N} \tau_i \mathbf{w}_i
    \label{eq:target_barycenter}
\end{equation}
\end{minipage}
\begin{equation}
    \mathbf{z}^{(t+1)} = \mathbf{z}^{(t)} + \eta \left( \mathbf{z}_{target}^{(t)} - \mathbf{z}^{(t)} \right)
    \label{eq:vector_flow}
\end{equation}

\subsection{Gradient Ascent Strategy}

In the context of a Guided VAE, the gradient ascent strategy actively searches for a counterfactual path. Starting from a starting latent representation, the goal is to discover the minimal feature changes required to transition into a winning state. To ensure the generated counterfactuals remain realistic and do not exploit adversarial blind spots in the classifier, the trajectory is explicitly regularized by the data manifold. \textbf{Objective Formulation:}~The optimization seeks to iteratively adjust the latent vector $\mathbf{z}$ to maximize an opponent-aware classification score $S(\mathbf{z})$ (e.g., the logit of winning against a specific opponent $\mathbf{z}_{opp}$), constrained by a manifold density penalty $D(\mathbf{z})$. The density is modeled using a fully differentiable Gaussian Kernel Density Estimate (KDE) evaluated over the reference dataset of known winning latents $\mathcal{Z}_{win} = \{\mathbf{w}_1, \dots, \mathbf{w}_N\}$ with bandwidth $h$, as seen in \autoref{eq:kde_density}. The total combined gradient at step $t$ merges the direction that increases the likelihood of winning with the direction that points toward denser, realistic regions of the latent space, as shown in \autoref{eq:total_gradient}, where $\lambda$ represents the weighting of the density prior (\texttt{density\_weight}). \textbf{Momentum-Based Optimization Update:}~To traverse the disentangled latent space smoothly and avoid local minima, the latent vector is updated using gradient ascent with momentum. Let $\alpha$ be the learning rate and $\beta$ be the momentum factor. The velocity $\mathbf{v}$ and position $\mathbf{z}$ are updated as seen in \autoref{eq:velocity_update}, and \autoref{eq:position_update}. \textbf{Convergence and Resampling:}~This iterative process continues until the predicted probability of the winning class exceeds a specified \texttt{convergence\_threshold} $\tau$ as in \autoref{eq:convergence_condition}. Because the number of optimization steps $T$ required to reach this threshold is variable, the resulting sequence $\{ \mathbf{z}^{(0)}, \mathbf{z}^{(1)}, \dots, \mathbf{z}^{(T)} \}$ is evenly resampled to extract exactly $n$ waypoints. This final trajectory is then output as the standardized path matrix $\mathbf{P} \in \mathbb{R}^{n \times d}$, representing a smooth, realistic counterfactual feature transition.

\vspace{\abovedisplayskip}
\noindent
\begin{minipage}{0.5\textwidth}
\begin{equation}
    D(\mathbf{z}) = \log \sum_{i=1}^{N} \exp \left( - \frac{1}{2h^2} \left\| \mathbf{z} - \mathbf{w}_i \right\|^2 \right)
    \label{eq:kde_density}
\end{equation}
\end{minipage}\begin{minipage}{0.5\textwidth}
\begin{equation}
    \mathbf{g}^{(t)}_{total} = \nabla_{\mathbf{z}} S(\mathbf{z}^{(t)}) + \lambda \nabla_{\mathbf{z}} D(\mathbf{z}^{(t)})
    \label{eq:total_gradient}
\end{equation}
\end{minipage}
\vspace{\belowdisplayskip}
\noindent
\begin{minipage}{0.5\textwidth}
\begin{equation}
    \mathbf{v}^{(t+1)} = \beta \mathbf{v}^{(t)} + \alpha \mathbf{g}^{(t)}_{total}
    \label{eq:velocity_update}
\end{equation}
\end{minipage}\begin{minipage}{0.5\textwidth}
\begin{equation}
    \mathbf{z}^{(t+1)} = \mathbf{z}^{(t)} + \mathbf{v}^{(t+1)}
    \label{eq:position_update}
\end{equation}
\end{minipage}
\begin{equation}
    P(\text{win} \mid \mathbf{z}^{(t)}, \mathbf{z}_{opp}) \ge \tau
    \label{eq:convergence_condition}
\end{equation}

\subsection{Neural Flow Strategy}

The neural flow strategy utilizes continuous normalizing flows via an Optimal Transport (OT) Flow Matching framework. Rather than relying on simple geometric interpolations or local gradient steps, this approach trains a neural network to learn a global velocity field. This field models the continuous optimal transport of probability mass from the "losing" to the "winning" latent distribution. \textbf{Velocity Field Training and OT Pairing:}~A Multi-Layer Perceptron (MLP) acts as a time-conditioned velocity field $\mathbf{v}_\theta(\mathbf{z}, t)$, parameterized by weights $\theta$. During training, mini-batches of losing latents $\mathbf{Z}_0$ and winning latents $\mathbf{Z}_1$ are extracted. To ensure the network learns the most efficient, non-crossing paths between these distributions, the samples are dynamically paired using exact Earth Mover's Distance (EMD) based on squared Euclidean distance. For each $\mathbf{z}_0$, an optimal $\mathbf{z}_{1, \text{paired}}$ is identified. At a uniformly sampled time $t \in [0, 1]$, the intermediate state is defined by linear interpolation as shown in \autoref{eq:interp_state}. The network is then trained to predict the constant-velocity vector between paired samples by minimizing the Mean Squared Error (MSE), as shown in \autoref{eq:mse_loss}. \textbf{Trajectory Integration (Euler Method):}~To generate a counterfactual path during inference, a starting (losing) latent $\mathbf{z}_{start}$ is integrated through the learned velocity field from $t=0$ to $t_{max} = 1.0$. Using a discrete number of integration steps $N_{steps}$, the time increment is $\Delta t = \frac{1.0}{N_{steps}}$. The latent position is updated iteratively using the Euler method seen in \autoref{eq:euler}, where the initial condition is $\mathbf{z}^{(0)} = \mathbf{z}_{start}$. This produces a smooth flow along the learned data manifold. \textbf{Classifier Guidance (Optional):}~To explicitly steer the trajectory toward regions with a higher probability of winning against a specific opponent, optional classifier guidance can be injected into the Euler integration. At each step, the gradient of the probability score $S(\mathbf{z})$ is computed. To ensure the guidance scale remains a consistent fraction of the flow step size, regardless of the raw gradient's magnitude, the gradient is normalized to a unit vector. The position is updated as shown in \autoref{eq:classifier_guidance}, where $\gamma$ (\texttt{guidance\_scale}) dictates how strongly the path is pulled toward the classifier's optimal regions, and $\epsilon$ prevents division by zero.

\vspace{\abovedisplayskip}
\noindent
\begin{minipage}{0.4\textwidth}
\begin{equation}
    \mathbf{z}_t = (1 - t) \mathbf{z}_0 + t \mathbf{z}_{1, \text{paired}}
    \label{eq:interp_state}
\end{equation}
\end{minipage}\begin{minipage}{0.6\textwidth}
\begin{equation}
    \mathcal{L}(\theta) = \mathbb{E}_{t, \mathbf{z}_0, \mathbf{z}_1} \left[ \left\| \mathbf{v}_\theta(\mathbf{z}_t, t) - (\mathbf{z}_{1, \text{paired}} - \mathbf{z}_0) \right\|_2^2 \right]
    \label{eq:mse_loss}
\end{equation}
\end{minipage}

\noindent
\begin{minipage}{0.3\textwidth}
\begin{equation}
    \mathbf{z}^{(t+1)} = \mathbf{z}^{(t)} + \mathbf{v}_\theta(\mathbf{z}^{(t)}, t) \Delta t
    \label{eq:euler}
\end{equation}
\end{minipage}\begin{minipage}{0.7\textwidth}
\begin{equation}
    \mathbf{z}^{(t+1)} = \mathbf{z}^{(t)} + \Delta t \left( \mathbf{v}_\theta(\mathbf{z}^{(t)}, t) + \gamma \frac{\nabla_\mathbf{z} S(\mathbf{z}^{(t)})}{\|\nabla_\mathbf{z} S(\mathbf{z}^{(t)})\|_2 + \epsilon} \right)
    \label{eq:classifier_guidance}
\end{equation}
\end{minipage}
\vspace{\belowdisplayskip}
 
\section{Experiments}
\label{sec:results}

\textbf{Model Performance:} To evaluate the quality of the trained Guided VAE, we assess both its reconstruction capabilities and the effectiveness of the latent space separation. The model's performance on the held-out test set is summarized using several key metrics as seen in \autoref{tab:model-evaluation}. 

\begin{table}[htbp]
\centering
\caption{GuidedVAE Test-Set Evaluation}
\label{tab:model-evaluation}
\begin{tabular}{lrr}
\toprule
\textbf{Metric} & \textbf{SC2EGSet} & \textbf{OOD Data} \\
\midrule
\multicolumn{3}{l}{\textit{VAE}} \\
    MSE (original scale) & 430095.6 & 519342.7 \\
    MSE (normalised scale) & 0.5830 & 1.6619 \\
    KL Divergence & 7.3601 & 7.6009 \\
\addlinespace
\multicolumn{3}{l}{\textit{Classifier}} \\
    Accuracy (\%) & 98.76 & 97.11 \\
    ROC-AUC & 0.9991 & 0.9926 \\
    F1 Score & 0.9881 & 0.9714 \\
    Brier Score & 0.0090 & 0.0229 \\
\bottomrule
\end{tabular}
\end{table}

Evaluation of the GuidedVAE demonstrates strong generative and reconstruction fidelity, with the MSE and KL divergence confirming accurate game-state reconstruction from a well-regularised latent space. Furthermore, evaluation of the predictive guidance imposed on the first latent dimension—measured via test accuracy, ROC-AUC, and Brier score—indicates robust classification performance and highly calibrated win probabilities. Overall, the model successfully balances precise feature reconstruction with meaningful latent disentanglement, establishing a reliable foundation for generating counterfactual improvement trajectories. \textbf{Conterfactual Paths:} To evaluate the performance of the final model against our main goal of latent space traversal generating counterfactual ``improvement trajectory'', we conduct a comparative assessment of all path generation strategies in \autoref{tab:cross-dataset}. Where the success rate is the fraction of samples for which the path reaches $P(\text{win}) \geq 0.5$ at any waypoint along the path. Crossover $\alpha$ is the position along the path at which $P(\text{win})$ first crosses 0.5, where $\alpha = 0$ is the start and $\alpha = 1$ is the end. Reported only for successful runs, $\bm{\Delta}$P(win) is the absolute gain in predicted win probability from path start to path end, i.e.\ $P(\text{win})_{\text{end}} - P(\text{win})_{\text{start}}$., AUC is the area under the $P(\text{win})$ curve over $\alpha \in [0,1]$. Monotonicity is the fraction of consecutive waypoint pairs for which $P(\text{win})$ is non-decreasing. A value of 1.0 means $P(\text{win})$ increases or stays flat at every step. Lower values indicate oscillation or regression along the path. Finally, the nearest-win distance is the Euclidean distance in the supervised latent subspace between the path endpoint and the closest winning latent vector in the training set. While Gradient Ascent achieves a nominally perfect success rate ($1.000$), further inspection of the secondary metrics suggests this performance is largely driven by adversarial off-manifold drift. Compared to geometrically grounded strategies such as Optimal Transport, Gradient Ascent tends to explore low-density regions of the latent space. This is directly evidenced by a heavily inflated maximum latent norm ($\max \|\mathbf{z}\| = 3.95 \pm 1.74$, compared to just $2.03 \pm 0.58$ for Optimal Transport) and a severely degraded path KDE density ($-4.63 \pm 1.40$ vs. $-2.80 \pm 0.50$). Furthermore, its significantly reduced monotonicity ($0.727 \pm 0.227$ vs. $0.998 \pm 0.041$) and higher nearest-win distance ($0.15 \pm 0.12$ vs. $0.06 \pm 0.04$) indicate erratic, unconstrained traversal rather than smooth semantic interpolation. Consequently, the representations generated by this strategy are likely to exploit classifier blind spots and warrant much closer examination, see~\appendixref{sec:additional_results}. We hypothesize that these failure modes could be addressed in future work by imposing stricter regularization constraints to firmly anchor the trajectory to the learned data prior.

\begin{table}[htbp]
\centering
\caption{Cross-Dataset Comparison of Path-Charting Strategies}
\label{tab:cross-dataset}
\begin{tabular}{llccc}
\toprule
\textbf{Method} & \textbf{Metric} & \textbf{SC2EGSet} & \textbf{OOD Data} & $\bm{\Delta}$ Dataset \\
\midrule
Linear (centroid) & Success rate & 0.834 & 0.715 & -0.118 \\
 & Crossover $\alpha$ & 0.659 $\pm$ 0.202 & 0.704 $\pm$ 0.236 & +0.045 \\
 & $\bm{\Delta}$P(win) & 0.813 $\pm$ 0.320 & 0.681 $\pm$ 0.399 & -0.132 \\
 & AUC & 0.310 $\pm$ 0.221 & 0.234 $\pm$ 0.236 & -0.076 \\
 & Monotonicity & 0.997 $\pm$ 0.041 & 0.977 $\pm$ 0.093 & -0.020 \\
 & Nearest-win dist. & 0.06 $\pm$ 0.04 & 0.05 $\pm$ 0.03 & -0.006 \\
\addlinespace
Linear (k-NN) & Success rate & 0.845 & 0.543 & -0.303 \\
 & Crossover $\alpha$ & 0.688 $\pm$ 0.190 & 0.700 $\pm$ 0.250 & +0.012 \\
 & $\bm{\Delta}$P(win) & 0.824 $\pm$ 0.310 & 0.510 $\pm$ 0.443 & -0.314 \\
 & AUC & 0.290 $\pm$ 0.206 & 0.182 $\pm$ 0.234 & -0.108 \\
 & Monotonicity & 0.998 $\pm$ 0.027 & 0.966 $\pm$ 0.128 & -0.032 \\
 & Nearest-win dist. & 0.05 $\pm$ 0.05 & 0.06 $\pm$ 0.04 & +0.005 \\
\addlinespace
Optimal Transport & Success rate & 0.837 & 0.720 & -0.117 \\
 & Crossover $\alpha$ & 0.120 $\pm$ 0.065 & 0.140 $\pm$ 0.084 & +0.020 \\
 & $\bm{\Delta}$P(win) & 0.819 $\pm$ 0.315 & 0.682 $\pm$ 0.397 & -0.137 \\
 & AUC & 0.752 $\pm$ 0.301 & 0.629 $\pm$ 0.364 & -0.123 \\
 & Monotonicity & 0.998 $\pm$ 0.041 & 0.995 $\pm$ 0.059 & -0.003 \\
 & Nearest-win dist. & 0.06 $\pm$ 0.04 & 0.05 $\pm$ 0.03 & -0.005 \\
\addlinespace
Neural Flow & Success rate & 0.931 & 0.731 & -0.200 \\
 & Crossover $\alpha$ & 0.524 $\pm$ 0.230 & 0.494 $\pm$ 0.279 & -0.030 \\
 & $\bm{\Delta}$P(win) & 0.915 $\pm$ 0.241 & 0.701 $\pm$ 0.427 & -0.213 \\
 & AUC & 0.468 $\pm$ 0.253 & 0.390 $\pm$ 0.332 & -0.078 \\
 & Monotonicity & 0.999 $\pm$ 0.013 & 0.996 $\pm$ 0.025 & -0.002 \\
 & Nearest-win dist. & 0.18 $\pm$ 0.16 & 0.20 $\pm$ 0.19 & +0.024 \\
\addlinespace
Gradient Ascent & Success rate & 1.000 & 0.998 & -0.002 \\
 & Crossover $\alpha$ & 0.486 $\pm$ 0.367 & 0.513 $\pm$ 0.337 & +0.027 \\
 & $\bm{\Delta}$P(win) & 0.986 $\pm$ 0.076 & 0.969 $\pm$ 0.139 & -0.017 \\
 & AUC & 0.545 $\pm$ 0.355 & 0.517 $\pm$ 0.321 & -0.029 \\
 & Monotonicity & 0.727 $\pm$ 0.227 & 0.749 $\pm$ 0.209 & +0.021 \\
 & Nearest-win dist. & 0.15 $\pm$ 0.12 & 0.15 $\pm$ 0.14 & +0.005 \\
\addlinespace
\bottomrule
\end{tabular}
\\[0.5em]
\raggedright

\end{table}
 
\begin{figure}[H]
  \centering
  \includegraphics[width=\textwidth]{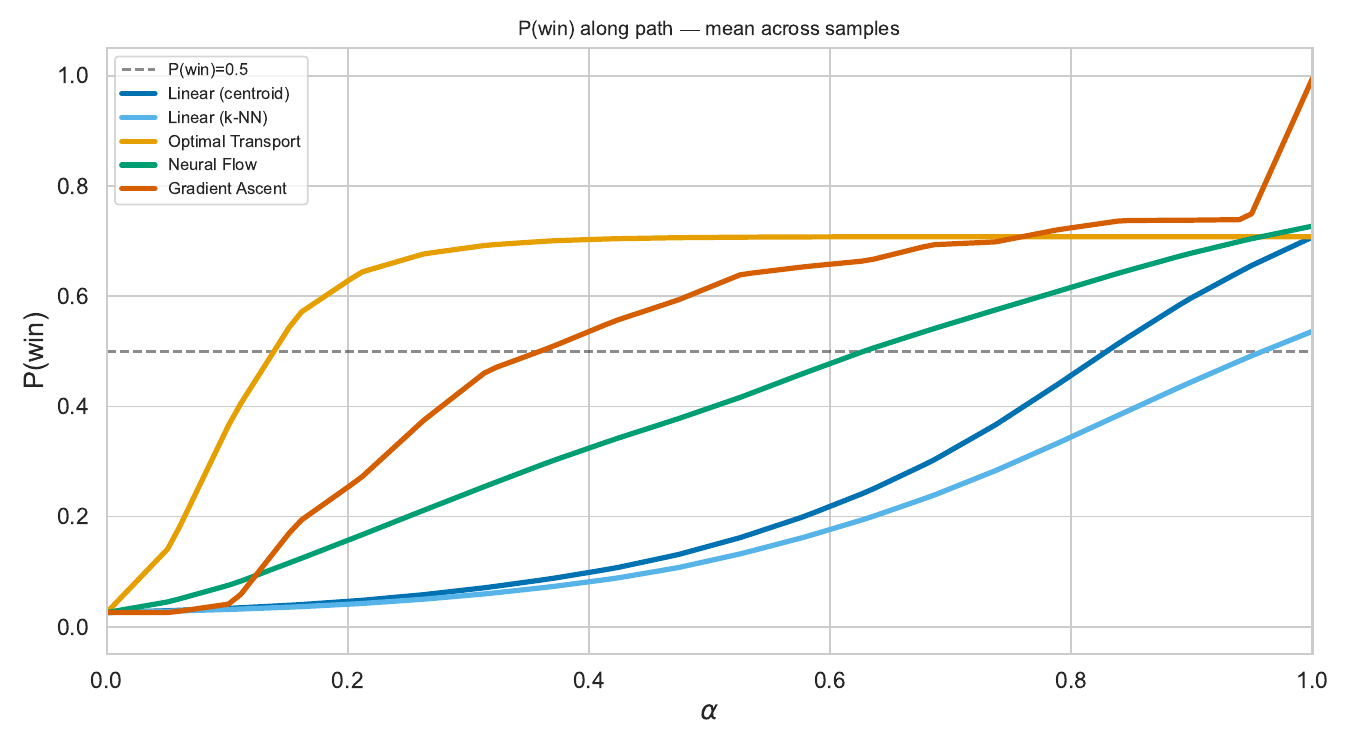}
  \caption{Mean OOD data $P(\text{win})$ performance of generated paths progress in the latent space.}
  \label{fig:per_strategy_win_curves}
\end{figure}

The experimental results reveal several key insights regarding the trade-offs between path reliability and quality:
\begin{enumerate*}[label=(\arabic*)]
    \item \textbf{Reliability and Success Rates:} Gradient Ascent emerges as the strategy maintaining a near-perfect success rate on both the \textbf{SC2EGSet} (1.000) and \textbf{OOD Data} (0.998). In contrast, the Linear (k-NN) baseline exhibits significant fragility under distributional shift, with success rates dropping by over 30\% ($\Delta = -0.302$).
    \item \textbf{Path Efficiency and Crossover:} Optimal Transport (OT) demonstrates superior efficiency in trajectory charting. As shown by the \textbf{Crossover $\alpha$} ($0.120 \pm 0.065$), OT-generated paths transition to a winning state much earlier than Linear methods ($\alpha \approx 0.65$). Furthermore, OT achieves the highest \textbf{AUC} ($0.752 \pm 0.301$), suggesting it identifies more direct routes through the latent space.
    \item \textbf{The Success-Monotonicity Trade-off :} A clear divergence exists between raw success and path smoothness. While Gradient Ascent is the most successful, it records the lowest \textbf{Monotonicity} ($0.727 \pm 0.227$), indicating more erratic trajectories. Conversely, Neural Flow and Optimal Transport maintain near-perfect monotonicity ($>0.99$) even on OOD data, providing highly stable and interpretable transitions.
\end{enumerate*}
The substantial variance in $\Delta P(\text{win})$ across the linear baselines further suggests that simple interpolation is insufficient to capture the model's complex decision boundaries. In contrast, neural and transport-based methods provide more consistent counterfactual evidence, with a mean $P(\text{win})$ along the counterfactual path shown in~\autoref{fig:per_strategy_win_curves}, and directly showcase some of the aforementioned trade-offs on OOD data. Further model interpretability is covered in~\appendixref{sec:additional_results}.

\section{Limitations and Future Research}
\label{sec:limitations}

\textbf{Limitations:} Despite the promising results, our approach has several limitations that should be acknowledged. First, we train our model on a dataset of games spanning multiple years. We do not explicitly account for the game updates. Additionally, we do not encode the players' in-game race information in any way. By design, our model is incapable of providing feedback directed towards specific in-game actions and environment configurations. The model feedback is additionally constrained by the dataset choice; tournament gameplay samples can be seen as a very specific subset of all of the games. Our model does not directly convey a more granular approach of jumping between leagues when leveraging its feedback. Finally, the model and method parameters were not verified with human participants, and aside from expert input from known professional players, we were unable to set up a human-in-the-loop type of experiment. \textbf{Future Research:} We hope that by extending research efforts in representational learning geared towards providing feedback, we can inspire others to prepare end-to-end feedback generation. Automating ways to improve humans based on deep generative solutions and other computational means. In the future, we wish to address most of the concerns raised above. Moving towards optimizing human performance jointly with actions against an environment sounds incredibly exciting, with the potential to uncover environment configurations that promote positive training or learning outcomes. Given the recent advancements and rapid adoption of AI systems, creating models that provide data-driven, actionable feedback is crucial. 
\section{Conclusion and Summary}
\label{sec:summary}

We have demonstrated the computational feasibility and the potential for developing many practical solutions for providing feedback based on learned representations. Additionally, simplistic methods, while appealing, have drawbacks that become more evident when dealing with a more advanced nonlinear model. Finally, we have accomplished our goal of bridging the gap between representational learning and generating feedback by extracting actionable, counterfactual improvement trajectories from the latent space, effectively shifting the analytical paradigm from predicting game outcomes to providing players with tangible guidance on what they should do differently. Based on our ongoing discussions with esports professionals, our solution is proving to be a real asset for future feedback systems.

\bibliographystyle{IEEEtranN}
\bibliography{IEEEabrv,sources.bib}

\appendix

\renewcommand{\sectionautorefname}{Appendix}

\section{StarCraft II Game Description}
\label{sec:sc2_description}

Before diving deeper it is important to state the rules governing StarCraft~II competitive gameplay. The game contains three main ``races'' of choice for the players. Each race is differentiated by unit mechanics therefore forcing certain playstyles. In most cases the game is played in the format of player versus player (PvP), one versus one (1 vs 1). The ultimate goal for competitors is to destroy all of the opponents' structures, or to force their counterpart to resign. In many cases tournaments have varying stages, such as the initial group stage, and a subsequent knockout bracket stage. Depending on the tournament format in most cases the group stages are played out as best of one (Bo1), best of two (Bo2), or best of three (Bo3) matches. Finally, the knockout bracket features Bo3, best of five (Bo5), and best of seven (Bo7) matches.

\section{Model Details}
\label{sec:model_details}

\paragraph{Training Objective:} The VAE is trained with the standard reconstruction and KL divergence objective. The reconstruction term is mean squared error between the original feature vector and the decoded feature vector. The KL term regularizes the posterior distribution toward a unit Gaussian prior. The VAE loss is shown in \autoref{eq:vae_loss} where $\mathcal{L}_{\mathrm{MSE}}$ is the summed mean squared error between the input feature vector and its reconstruction, and $\mathcal{L}_{\mathrm{KL}}$ regularizes the approximate posterior toward a unit Gaussian prior. To guide the latent space, we add a binary cross-entropy loss on the prediction produced from the first latent dimension. The main Guided VAE update is seen on \autoref{eq:main_loss}, where $\lambda_{\mathrm{cls}}$ controls the strength of supervision and encourages outcome-relevant information to be concentrated in the guided latent dimension.

\noindent
\begin{minipage}{0.5\textwidth}
\begin{equation}
    \mathcal{L}_{\mathrm{VAE}} = \mathcal{L}_{\mathrm{MSE}} + \mathcal{L}_{\mathrm{KL}},
    \label{eq:vae_loss}
\end{equation}
\end{minipage}\begin{minipage}{0.5\textwidth}
\begin{equation}
    \mathcal{L}_{\mathrm{main}} = \mathcal{L}_{\mathrm{VAE}} + \lambda_{\mathrm{cls}} \mathcal{L}_{\mathrm{BCE}},
    \label{eq:main_loss}
\end{equation}
\end{minipage}
\vspace{\belowdisplayskip}

Aside from the typical structure of a Guided VAE model we clamp the log variance output of the encoder to the interval $[-20, 2]$ before the reparameterization step. This hard bound constrains the effective standard deviation to the range $[\approx 4.5 \times 10^{-5},\ \approx 2.7]$, preventing two sources of numerical instability: a near-zero variance, which causes the KL divergence term in the evidence lower bound (ELBO) to diverge and produces exploding gradients; and an excessively large variance, which overwhelms the mean and injects too much noise into the decoder input, destabilizing reconstruction.  These dimensions are used to predict the game outcome. The point of this is to force at least one direction of the latent space to be directly related to winning and losing. 

\section{Hardware}
\label{sec:hardware}

\subsection{Hardware and Computational Requirements}

All experiments were conducted on a high-performance workstation using consumer-grade hardware. The specific configuration of the system components is detailed in Table \ref{tab:hardware}.

\begin{table}[htbp]
\centering
\caption{Hardware specifications used for all experimental runs and hyperparameter sweeps.}
\label{tab:hardware}
\begin{tabular}{@{}ll@{}}
\toprule
\textbf{Component} & \textbf{Specification} \\
\midrule
CPU & AMD Ryzen 9 9950X (16-Core, 32-Thread) \\
GPU & NVIDIA GeForce RTX 5090 \\
Memory (RAM) & 128 GB DDR5 \\
\bottomrule
\end{tabular}
\end{table}

The model architectures and optimization strategies were designed for high efficiency. Consequently, the majority of individual experimental runs, including those within the Guided-VAE hyperparameter sweep and path generation evaluations, were completed in under 5 minutes.

\section{Hyperparameter Search}
\label{sec:hparams}

\subsection{Guided-VAE Hyperparameter Search}

\subsubsection{Search Space}

The search space for the Guided-VAE is defined by a set of hierarchical constraints to ensure a valid bottleneck architecture. Let $\mathcal{W} = \{w_0, w_1, \dots, w_m\}$ denote the set of available layer widths in ascending order.

\subsubsubsection{Architecture Constraints}

The encoder configuration is sampled via two primary parameters: the number of layers $n \in \{2, 3, 4\}$ and a categorical start width $w_{start} \in \mathcal{W}$. To guarantee that the resulting hidden dimensions $\mathbf{H}_{enc}$ are strictly decreasing, we calculate the actual starting index $i$ as:
\begin{equation}
    i = \max(n - 1, \text{index}(w_{start}))
\end{equation}
The sequence of hidden dimensions is then defined as:
\begin{equation}
    \mathbf{H}_{enc} = \{ w_{i-j} \}_{j=0}^{n-1}
\end{equation}

The latent dimensionality $n_z$ is coupled to the final encoder width $h_{last} \in \mathbf{H}_{enc}$ through a fraction $f_z \in \{0.25, 0.5, 1.0\}$, constrained by a minimum floor:
\begin{equation}
    n_z = \max(8, \lfloor h_{last} \cdot f_z \rfloor)
\end{equation}

\subsubsubsection{Optimization Parameters}
The remaining parameters are sampled according to the following distributions:
\begin{itemize}
    \item \textbf{Learning Rates:} $\eta_{vae}, \eta_{cls} \sim \text{LogUniform}(10^{-5}, 10^{-3})$
    \item \textbf{Weight Decays:} $\lambda_{vae}, \lambda_{cls} \sim \text{LogUniform}(10^{-6}, 10^{-3})$
    \item \textbf{Classification Weight:} $\alpha \sim \text{LogUniform}(1, 250)$
    \item \textbf{Supervised Dimensions:} $n_s \in \{1, 2, 4\}$
\end{itemize}

\subsubsection{Hyperparameter Run Configuration}

To find the best hyperparameters for our training, we ran the search using Ray (https://www.ray.io/) and Optuna (https://optuna.org/). Upon execution, we have decided on 150 total runs. The objective function, $\mathcal{O}_{HPO}$, is defined as a weighted scalar sum of validation metrics logged during the training of the Guided VAE model. Formally, the minimization objective is expressed as:

\begin{equation}
\min_{\theta} \mathcal{O}_{HPO} = \sum_{i \in \mathcal{M}} w_i \cdot \mathcal{L}_i(\theta)
\end{equation}

where $\mathcal{M}$ denotes the set of validation metrics, $\mathcal{L}_i$ is the value of the $i$-th metric, and $w_i$ is the user-defined weight for that metric. In the configuration utilized for this sweep, the objective was set to equally weight the reconstruction and classification components:

\begin{equation}
\mathcal{O}_{HPO} = w_{\text{vae}} \mathcal{L}_{\text{val\_vae}} + w_{\text{cls}} \mathcal{L}_{\text{val\_cls}}
\end{equation}

Given our configuration parameters $w_{\text{vae}} = 0.5$ and $w_{\text{cls}} = 0.5$, the final objective function simplifies to:

\begin{equation}
\mathcal{O}_{HPO} = 0.5(\mathcal{L}_{\text{val\_vae}}) + 0.5(\mathcal{L}_{\text{val\_cls}})
\end{equation}

\subsubsection{Guided VAE Final Hyperparameters}

\autoref{tab:guided_vae_hyperparams} contains the model we have selected for a best performing model.

\begin{table}[htbp]
\centering
\caption{Final hyperparameter values for the Guided-VAE model discovered via the Ray/Optuna optimization sweep.}
\label{tab:guided_vae_hyperparams}
\begin{tabular}{@{}llc@{}}
\toprule
\textbf{Category} & \textbf{Hyperparameter} & \textbf{Value} \\
\midrule
\textbf{Architecture} & Input Dimension & $196$ \\
& Encoder Hidden Dimensions ($\mathbf{H}_{enc}$) & $[32, 16]$ \\
& Latent Dimensionality ($n_z$) & $16$ \\
& Supervised Dimensions ($n_s$) & $4$ \\
\midrule
\textbf{Optimization} & VAE Learning Rate ($\eta_{vae}$) & $1.7725 \times 10^{-4}$ \\
& VAE Weight Decay ($\lambda_{vae}$) & $1.3597 \times 10^{-5}$ \\
& Classifier Learning Rate ($\eta_{cls}$) & $4.1398 \times 10^{-4}$ \\
& Classifier Weight Decay ($\lambda_{cls}$) & $4.5500 \times 10^{-6}$ \\
& Classification Weight ($\alpha$) & $1.2824$ \\
\bottomrule
\end{tabular}
\end{table}

\subsection{Path Generation Strategies: Hyperparameter Search}

\subsubsection{Search Space}

The search space for the latent space traversal is structured hierarchically, where the subset of active hyperparameters is conditioned on the chosen strategy $S$.

\paragraph{Linear Strategy}
The linear interpolation strategy relies on neighborhood density constraints:
\begin{itemize}
    \item \textbf{Nearest Neighbors ($k_{neighbors}$):} $k_{nb} \sim \text{DiscreteUniform}(3, 15)$
    \item \textbf{Opponent Constraints ($k_{opponents}$):} $k_{opp} \sim \text{DiscreteUniform}(10, 200)$
\end{itemize}

\paragraph{Gradient Ascent Strategy}
This strategy utilizes a density-based optimization approach with fixed steps $T=2000$ and a convergence threshold $\tau=0.95$:
\begin{itemize}
    \item \textbf{Learning Rate ($\eta$):} $\eta \sim \text{LogUniform}(10^{-4}, 0.1)$
    \item \textbf{Momentum ($\mu$):} $\mu \sim \text{Uniform}(0.0, 0.95)$
    \item \textbf{Density Weight ($w_{\rho}$):} $w_{\rho} \sim \text{Uniform}(0.0, 1.0)$
    \item \textbf{KDE Bandwidth ($h$):} $h \sim \text{Uniform}(0.1, 2.0)$
\end{itemize}

\paragraph{Optimal Transport Strategy}
The optimal transport strategy balances regularization and geometric constraints:
\begin{itemize}
    \item \textbf{Regularization ($\epsilon$):} $\epsilon \sim \text{LogUniform}(0.01, 0.5)$
    \item \textbf{Step Size ($\gamma$):} $\gamma \sim \text{Uniform}(0.05, 0.5)$
    \item \textbf{Opponent Constraints ($k_{opponents}$):} $k_{opp} \sim \text{DiscreteUniform}(10, 200)$
\end{itemize}

\paragraph{Neural Flow Strategy}
The neural flow strategy utilizes a fixed guidance scale for its transformation:
\begin{itemize}
    \item \textbf{Guidance Scale ($s$):} $s = 1.0$ (fixed)
\end{itemize}

\subsubsection{Hyperparameter Run Configuration}

The optimization of hyperparameters for the latent space traversal strategies was performed using the Optuna framework. For each of the strategy we executed $N=100$ independent trials. In each trial, the performance was evaluated by generating $n_{samples} = 1000$ latent paths.

Unlike the model training phase, the objective for path charting is a maximization task. The objective function $\mathcal{J}$ is defined as the mean performance of a specified evaluation metric $\mathcal{M}$: AUC across all generated samples:

\begin{equation}
\max_{\phi} \mathcal{J}(\phi) = \mathbb{E}_{s \sim \mathcal{S}(\phi)} [ \mathcal{M}(s) ]
\end{equation}

where $\phi$ represents the set of strategy-specific hyperparameters sampled from the search space, and $\mathcal{S}(\phi)$ denotes the distribution of paths generated under those parameters. 

\subsubsection{Path Generation Strategies: Final Hyperparameters}

\autoref{tab:hyperparameters_paths} contains the specific hyperparameters used for the final runs of our path generation strategies.

\begin{table}[htbp]
\centering
\caption{Hyperparameters used for the evaluated methods. Continuous values discovered via hyperparameter optimization are rounded to four decimal places.}
\label{tab:hyperparameters_paths}
\begin{tabular}{@{}llc@{}}
\toprule
\textbf{Method} & \textbf{Hyperparameter} & \textbf{Value} \\
\midrule
\textbf{Neural Flow} & Guidance Scale & $1.0$ \\
\midrule
\textbf{Gradient Ascent} & Steps & $2000$ \\
& Learning Rate (lr) & $0.0330$ \\
& Momentum & $0.8815$ \\
& Density Weight & $0.8444$ \\
& KDE Bandwidth & $0.6516$ \\
& Convergence Threshold & $0.95$ \\
\midrule
\textbf{Linear Centroid} & $k$ Neighbours & $14$ \\
& $k$ Opponents & $74$ \\
\midrule
\textbf{Linear Nearest} & $k$ Neighbours & $10$ \\
& $k$ Opponents & $11$ \\
\midrule
\textbf{Optimal Transport} & Regularization (reg) & $0.0887$ \\
& Step Size & $0.4999$ \\
& $k$ Opponents & $64$ \\
\bottomrule
\end{tabular}
\end{table}

\section{Additional Results}
\label{sec:additional_results}

\begin{table}[htbp]
\centering
\caption{Cross-Dataset Comparison of Path-Charting Strategies with all of the computed metrics.}
\label{tab:cross-dataset-full}
\begin{tabular}{llccc}
\toprule
\textbf{Method} & \textbf{Metric} & \textbf{SC2EGSet} & \textbf{OOD Data} & $\bm{\Delta}$ Dataset \\
\midrule
Linear (centroid) & Success rate & 0.834 & 0.715 & -0.118 \\
 & Crossover $\alpha$ & 0.659 $\pm$ 0.202 & 0.704 $\pm$ 0.236 & +0.045 \\
 & $\bm{\Delta}$P(win) & 0.813 $\pm$ 0.320 & 0.681 $\pm$ 0.399 & -0.132 \\
 & AUC & 0.310 $\pm$ 0.221 & 0.234 $\pm$ 0.236 & -0.076 \\
 & Monotonicity & 0.997 $\pm$ 0.041 & 0.977 $\pm$ 0.093 & -0.020 \\
 & Nearest-win dist. & 0.06 $\pm$ 0.04 & 0.05 $\pm$ 0.03 & -0.006 \\
 & KDE density shift & 1.51 $\pm$ 1.11 & 1.74 $\pm$ 1.48 & +0.230 \\
 & Path KDE density & -3.12 $\pm$ 0.74 & -3.21 $\pm$ 0.81 & -0.096 \\
 & Cycle error & 0.218 $\pm$ 0.096 & 0.239 $\pm$ 0.096 & +0.021 \\
 & Max $\|\mathbf{z}\|$ & 2.02 $\pm$ 0.58 & 1.93 $\pm$ 0.51 & -0.092 \\
\addlinespace
Linear (k-NN) & Success rate & 0.845 & 0.543 & -0.303 \\
 & Crossover $\alpha$ & 0.688 $\pm$ 0.190 & 0.700 $\pm$ 0.250 & +0.012 \\
 & $\bm{\Delta}$P(win) & 0.824 $\pm$ 0.310 & 0.510 $\pm$ 0.443 & -0.314 \\
 & AUC & 0.290 $\pm$ 0.206 & 0.182 $\pm$ 0.234 & -0.108 \\
 & Monotonicity & 0.998 $\pm$ 0.027 & 0.966 $\pm$ 0.128 & -0.032 \\
 & Nearest-win dist. & 0.05 $\pm$ 0.05 & 0.06 $\pm$ 0.04 & +0.005 \\
 & KDE density shift & 1.43 $\pm$ 1.06 & 1.59 $\pm$ 1.40 & +0.162 \\
 & Path KDE density & -3.16 $\pm$ 0.82 & -3.35 $\pm$ 0.94 & -0.197 \\
 & Cycle error & 0.206 $\pm$ 0.087 & 0.235 $\pm$ 0.095 & +0.029 \\
 & Max $\|\mathbf{z}\|$ & 2.05 $\pm$ 0.63 & 1.87 $\pm$ 0.56 & -0.180 \\
\addlinespace
Optimal Transport & Success rate & 0.837 & 0.720 & -0.117 \\
 & Crossover $\alpha$ & 0.120 $\pm$ 0.065 & 0.140 $\pm$ 0.084 & +0.020 \\
 & $\bm{\Delta}$P(win) & 0.819 $\pm$ 0.315 & 0.682 $\pm$ 0.397 & -0.137 \\
 & AUC & 0.752 $\pm$ 0.301 & 0.629 $\pm$ 0.364 & -0.123 \\
 & Monotonicity & 0.998 $\pm$ 0.041 & 0.995 $\pm$ 0.059 & -0.003 \\
 & Nearest-win dist. & 0.06 $\pm$ 0.04 & 0.05 $\pm$ 0.03 & -0.005 \\
 & KDE density shift & 1.50 $\pm$ 1.11 & 1.74 $\pm$ 1.48 & +0.239 \\
 & Path KDE density & -2.80 $\pm$ 0.50 & -2.74 $\pm$ 0.40 & +0.060 \\
 & Cycle error & 0.242 $\pm$ 0.094 & 0.237 $\pm$ 0.113 & -0.005 \\
 & Max $\|\mathbf{z}\|$ & 2.03 $\pm$ 0.58 & 1.93 $\pm$ 0.51 & -0.103 \\
\addlinespace
Neural Flow & Success rate & 0.931 & 0.731 & -0.200 \\
 & Crossover $\alpha$ & 0.524 $\pm$ 0.230 & 0.494 $\pm$ 0.279 & -0.030 \\
 & $\bm{\Delta}$P(win) & 0.915 $\pm$ 0.241 & 0.701 $\pm$ 0.427 & -0.213 \\
 & AUC & 0.468 $\pm$ 0.253 & 0.390 $\pm$ 0.332 & -0.078 \\
 & Monotonicity & 0.999 $\pm$ 0.013 & 0.996 $\pm$ 0.025 & -0.002 \\
 & Nearest-win dist. & 0.18 $\pm$ 0.16 & 0.20 $\pm$ 0.19 & +0.024 \\
 & KDE density shift & 0.52 $\pm$ 1.78 & 0.79 $\pm$ 1.49 & +0.270 \\
 & Path KDE density & -3.43 $\pm$ 1.11 & -3.62 $\pm$ 1.37 & -0.192 \\
 & Cycle error & 0.294 $\pm$ 0.143 & 0.311 $\pm$ 0.151 & +0.016 \\
 & Max $\|\mathbf{z}\|$ & 2.54 $\pm$ 0.81 & 2.26 $\pm$ 0.61 & -0.285 \\
\addlinespace
Gradient Ascent & Success rate & 1.000 & 0.998 & -0.002 \\
 & Crossover $\alpha$ & 0.486 $\pm$ 0.367 & 0.513 $\pm$ 0.337 & +0.027 \\
 & $\bm{\Delta}$P(win) & 0.986 $\pm$ 0.076 & 0.969 $\pm$ 0.139 & -0.017 \\
 & AUC & 0.545 $\pm$ 0.355 & 0.517 $\pm$ 0.321 & -0.029 \\
 & Monotonicity & 0.727 $\pm$ 0.227 & 0.749 $\pm$ 0.209 & +0.021 \\
 & Nearest-win dist. & 0.15 $\pm$ 0.12 & 0.15 $\pm$ 0.14 & +0.005 \\
 & KDE density shift & 0.63 $\pm$ 1.40 & 0.78 $\pm$ 1.46 & +0.152 \\
 & Path KDE density & -4.63 $\pm$ 1.40 & -4.20 $\pm$ 1.25 & +0.430 \\
 & Cycle error & 0.406 $\pm$ 0.184 & 0.410 $\pm$ 0.178 & +0.004 \\
 & Max $\|\mathbf{z}\|$ & 3.95 $\pm$ 1.74 & 3.91 $\pm$ 1.56 & -0.039 \\
\addlinespace
\bottomrule
\end{tabular}
\\[0.5em]
\raggedright

\end{table}
 
\subsection{Model Interpretability}
\label{sec:interpretability}

\begin{figure}[H]
  \centering
  \includegraphics[width=\textwidth]{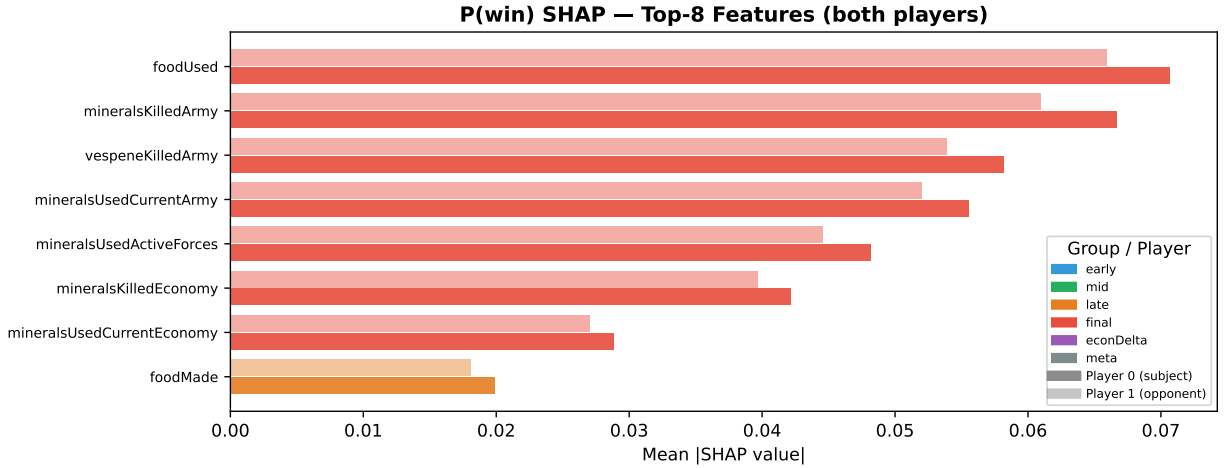}
  \caption{Mean absolute SHAP values for the top-8 input features of the GuidedVAE win-probability classifier $P(win)$.}
\label{fig:shap_bar}
\end{figure}

\begin{figure}[H]
  \centering
  \includegraphics[width=\textwidth]{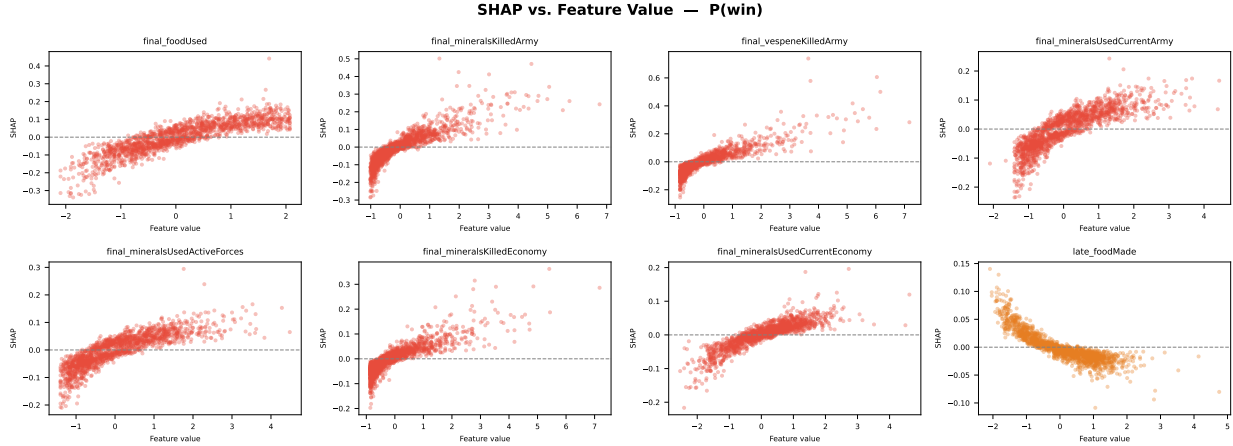}
  \caption{SHAP dependence plots for the top-8 features by mean $|SHAP|$ for $P(win)$.}
  \label{fig:shap_beeswarm}
\end{figure}

To verify that the Guided VAE's outcome classifier grounds its predictions in strategically meaningful features, we analyze its decision process using SHapley Additive exPlanations (SHAP) as shown in \autoref{fig:shap_bar} and \autoref{fig:shap_beeswarm}. SHAP values provide a unified measure of feature importance by attributing the log-odds of the predicted outcome to the individual input features. The mean absolute SHAP values highlight the top global contributors to the win-probability prediction, confirming that the model relies on core economic and macroscopic indicators rather than spurious correlations. Furthermore, the SHAP beeswarm plot reveals the distribution of these impacts across the test. It visualizes both the magnitude and the direction of the feature effects, illustrating how higher or lower values of specific features correlate with the predicted likelihood of winning. This transparency is crucial, as it ensures that the counterfactual paths generated by traversing the latent space correspond to interpretable, domain-consistent shifts in player behavior.

\section{Example Feedback Reports}
\label{sec:example_feedback}

\begin{figure}[H]
  \centering
  \includegraphics[width=\textwidth]{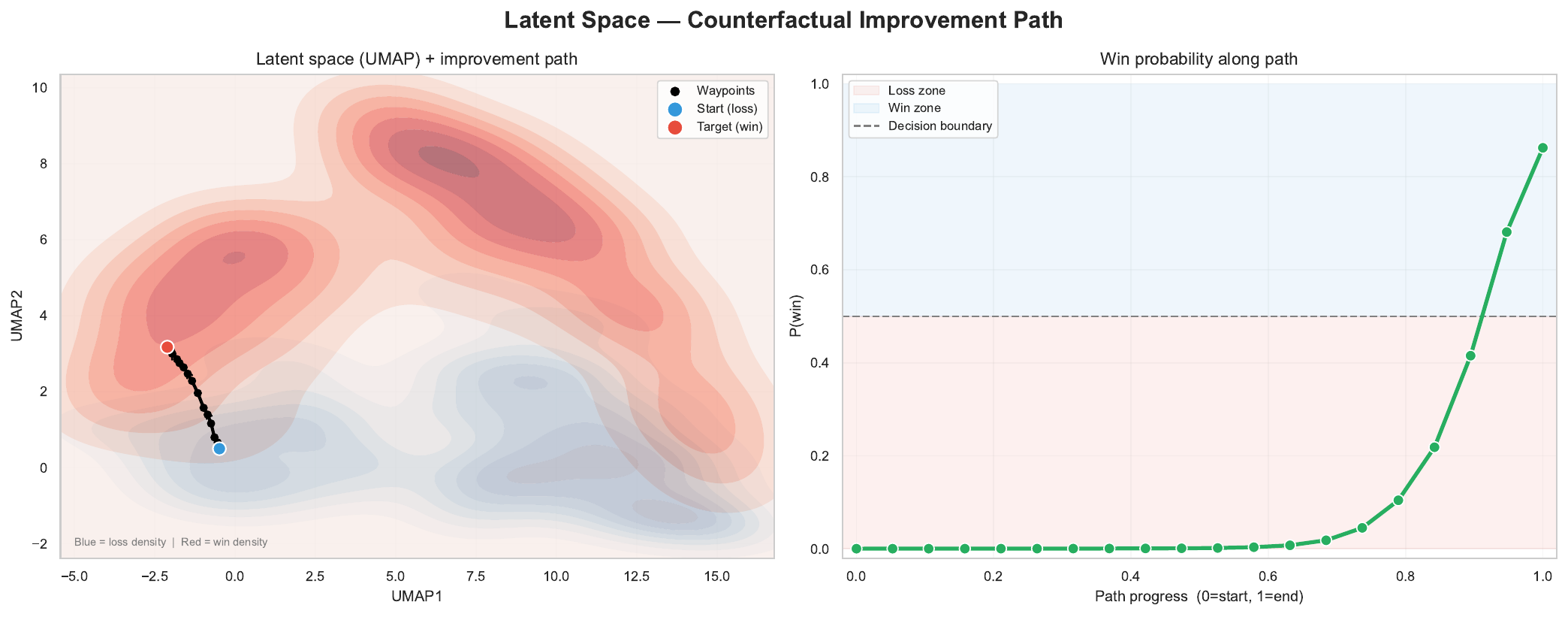}
  \caption{UMAP latent space projection with the counterfactual path shown.}
\label{fig:latent_space_density}
\end{figure}

\begin{figure}[H]
  \centering
  \includegraphics[width=\textwidth]{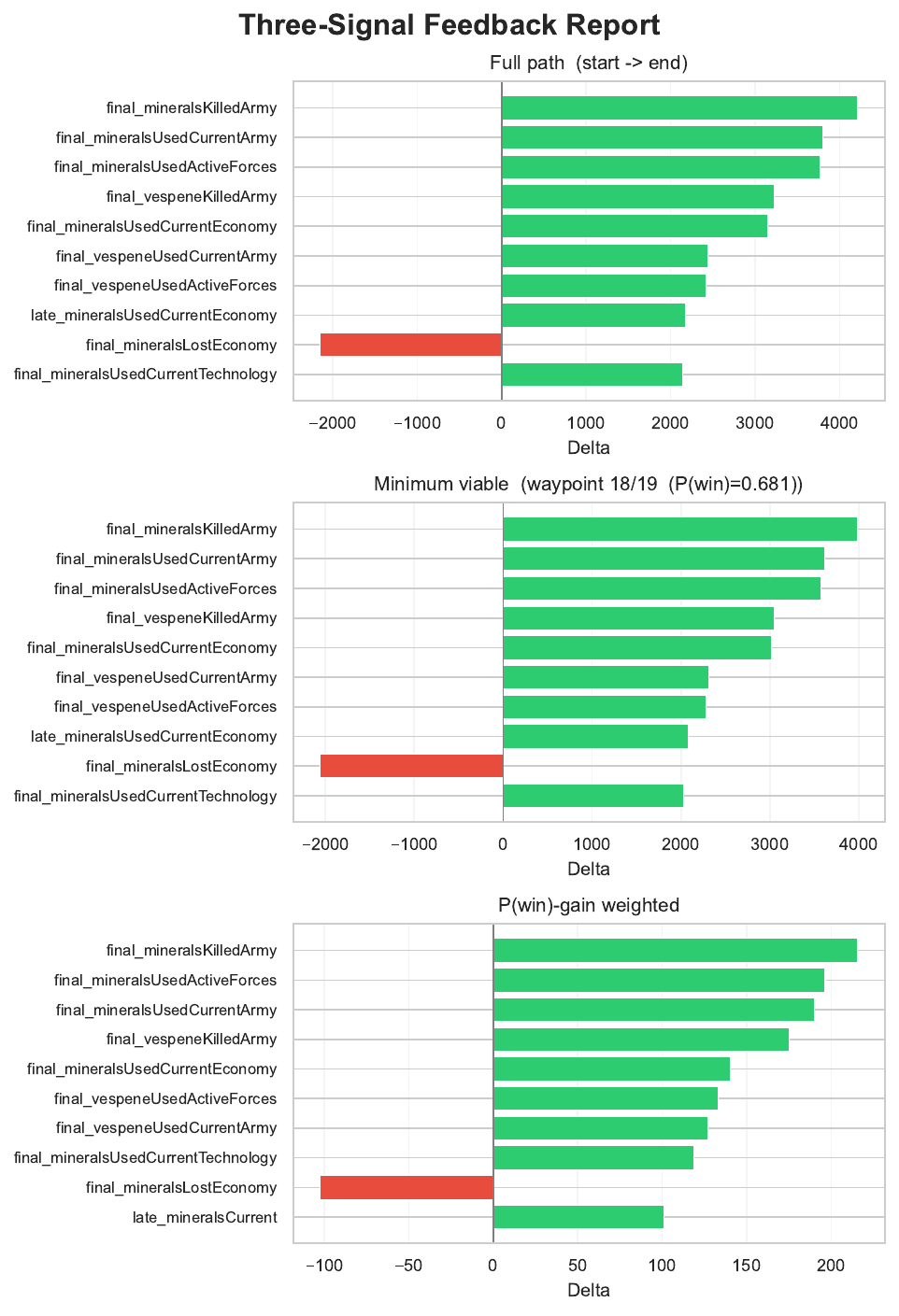}
  \caption{Feedback report with three distinct user interpretable signals.}
  \label{fig:three_signal}
\end{figure}

\autoref{fig:latent_space_density} and \autoref{fig:three_signal} are examples of the output, informing the user about the counteractual latent space path, and which parameters they should focus on.

\section{Data and Code Repositories}
\label{sec:appendix_data_code}

Anonymized version of our code and the pre-processed tensor data, as well as the model checkpoint are available at: \href{https://anonymous.4open.science/r/SC2\_LatentTrainer-1E5B/}{https://anonymous.4open.science/r/SC2\_LatentTrainer-1E5B/}. The original data repository is: \href{https://huggingface.co/datasets/Kaszanas/SC2EGSet}{https://huggingface.co/datasets/Kaszanas/SC2EGSet} 
\end{document}